\documentclass[twoside]{article}

\usepackage{PRIMEarxiv}

\usepackage[utf8]{inputenc} % allow utf-8 input
\usepackage[T1]{fontenc}    % use 8-bit T1 fonts
\usepackage{hyperref}       % hyperlinks
\usepackage{url}            % simple URL typesetting
\usepackage{booktabs}       % professional-quality tables
\usepackage{amsfonts}       % blackboard math symbols
\usepackage{nicefrac}       % compact symbols for 1/2, etc.
\usepackage{microtype}      % microtypography
\usepackage{lipsum}
\usepackage{fancyhdr}       % header
\usepackage{graphicx}       % graphics
\graphicspath{{media/}}     % organize your images and other figures under media/ folder
\usepackage{amsmath} 
\usepackage{float}

\usepackage{enumitem}
\usepackage{soul}
\usepackage{multirow}
\usepackage{multicol}
\usepackage{xcolor}
\usepackage{array}

\newcommand{\kt}[1]{{\color{black}#1}}
\newcommand{\rr}[1]{{\color{black}#1}} %% rochana used to be blue
\definecolor{darkgreen}{rgb}{0,0.7,0}

\newcommand{\g}[1]{{\textcolor{darkgreen}{#1}}}
\newcommand{\R}[1]{{\textcolor{red}{#1}}}

\title{[Re] Network Deconvolution}

\author{
  Rochana R. Obadage \\
  Old Dominion University  \\
  Norfolk, VA, USA \\
  \texttt{rochana@cs.odu.edu} \\
  \And
  Kumushini Thennakoon  \\
  Old Dominion University  \\
  Norfolk, VA, USA \\
  \texttt{kumushini@cs.odu.edu} \\
  \And
  Sarah M. Rajtmajer  \\
  IST, Pennsylvania State University  \\
  University Park, PA, USA \\
  \texttt{smr48@psu.edu} \\
  \And
  Jian Wu \\
  Old Dominion University  \\
  Norfolk, VA, USA \\
  \texttt{j1wu@odu.edu} \\
}

\begin{document}
\maketitle

\begin{abstract}
\kt{Our work aims to reproduce the set of findings published in "Network Deconvolution" by Ye et al. (2020)\cite{ye2020network}. That paper proposes an optimization technique for model training in convolutional neural networks. The proposed technique "network deconvolution" is used in convolutional neural networks to remove pixel-wise and channel-wise correlations before data is fed into each layer. In particular, we interrogate the validity of the authors' claim that using network deconvolution instead of batch normalization improves deep learning model performance. Our effort confirms the validity of this claim, successfully reproducing the results reported in Tables 1 and 2 of the original paper. Our study involved 367 unique experiments across multiple architectures, datasets, and hyper parameter configurations. For Table 1, while there were some minor deviations in accuracy when compared to the original values (within 10\%), the overall trend was consistent with the original study's findings when training the models with epochs 20 and 100. For Table 2, all 14 reproduced values were consistent with the original values. Additionally, we document the training and testing times for each architecture in Table 1 with 1, 20, and 100 epoch settings for both CIFAR-10 and CIFAR-100 datasets. We document the total execution times for Table 2 architectures with the ImageNet dataset. The data and software used for this reproducibility study are publicly available at \textit{\url{https://github.com/lamps-lab/rep-network-deconvolution}}}
\end{abstract}

% keywords can be removed

\section{Introduction}
\begin{figure}[H]
  \centering
    \setlength{\fboxsep}{2pt} % Adjust the padding
    \setlength{\fboxrule}{0.4pt} % Adjust the border width
    \fbox{\includegraphics[width=0.95\linewidth]{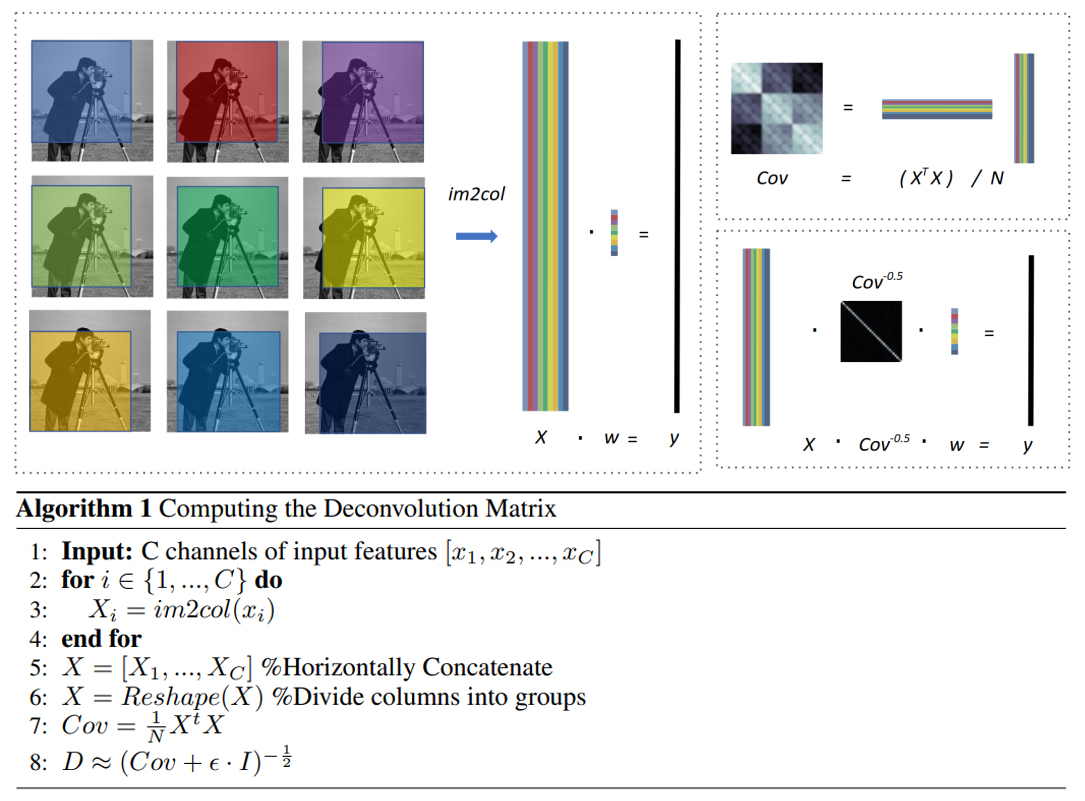}}
    % Adjust width as needed
    \caption{\textbf{Network deconvolution. Adopted from Figure 3 and Algorithm 1 in the original study \cite{ye2020network}}.}
    \label{fig:ND_operation}
\end{figure}

\kt{Batch normalization (BN) is a widely adopted technique in modern deep learning architectures, which has been shown to accelerate training and improve prediction performance \cite{BN-ioffe2015batch}. Recent studies have explored alternatives to BN to further improve model performance. One such approach is network deconvolution, proposed in 2020 by Ye et al. \cite{ye2020network} (hereafter, \textbf{original study}). 
As shown in the Figure \ref{fig:ND_operation}, the authors of the original study claim that network deconvolution enhances the training of convolutional neural networks (CNNs) by removing pixel-wise and channel-wise correlations in the input data. In CNNs, the blur effect occurs when an image is convolved with a Gaussian kernel or a similar smoothing filter. This process involves overlaying the kernel onto the image and performing a mathematical operation at each pixel. The blur effect which occurs due to the correlation of pixels reduces the clarity of object boundaries and features, making it difficult to accurately identify and localize objects.\\ }

% \begin{figure}[htbp]

 \rr{Per the original study, the architecture of network deconvolution involves replacing BN layers with deconvolution layers before a convolutional or fully-connected layer. This process entails calculating the covariance matrix of the input data, approximating its inverse square root, and then applying this transformation to decorrelate the data before feeding it into each subsequent layer (Figure \ref{fig:ND_operation}). 
 }

In the results reported in Table 1 of the original study \cite{ye2020network}, the authors evaluated their approach on 10 CNN architectures, including VGG-16, ResNet, PreAct-18, DenseNet-121, ResNeXt-29, MobileNet v2, DPN-92, PNASNetA, SENet-18, and EfficientNet. They used two benchmark datasets: CIFAR-10 \cite{cifar10} and CIFAR-100 \cite{cifar100}. Their results demonstrated improved performance compared with the baseline models using BN. In Table 2 of the original study, the authors compared VGG-11, ResNet-18, and DenseNet-121 architectures over BN and network deconvolution using the ImageNet \cite{deng2009imagenet} dataset to further validate the claims they obtained using CIFAR-10 and CIFAR-100.\\ 

Motivated by use cases including preserving spatial information \cite{spatial-information-preservation-cao2020sipmask}, reconstructing high-resolution features \cite{passarella2022-reconstructing}, improving localization accuracy \cite{localization-accuracy-gidaris2016locnet}, and image segmentation \cite{Image-segmentation-minaee2021image}, our study attempts to reproduce the results reported in the original paper with the most recent versions of software libraries. Therefore, our work used the same datasets and methods but different versions of the software packages adopted by the original paper. We define our work as a soft-reproducible study, to be distinct from a strict reproducible study, which adopts exactly the same datasets, and methods (including its software implementations). Soft-reproducibility studies circumvent many dependency compatibility obstacles are thus easier to implement. The results are still meaningful to understand the reliability of the original work.

\section{Citations of the Original Study from Scholarly Papers}

To analyze the scholarly reception of the original study, we reviewed 59 citation contexts from 31 citing papers extracted using the Semantic Scholar Graph API \cite{s2GA-Kinney2023TheSS}. The sentiment of each citation context was manually categorized as negative, positive, or neutral %based on hints on reproducibility in which \textbf{negative} means the context hints 
where \textbf{negative} was taken to indicate irreproducibility (e.g., unavailability of the cited paper’s data/code or unsuccessful attempts in reproducing), \textbf{positive} context suggests reproducibility (e.g., re-usage of the cited paper’s data/code or key concepts), and \textbf{neutral} means the context simply mentions (cites) the paper without offering information about its reproducibility. The classification process followed the same methodology as our previous study \cite{obadage2024-10.1145/3641525.3663628}. Of the 59 citation contexts, 13 were determined to be positive. The remaining 46 contexts were neutral. Importantly, there were no negative citation contexts. Two examples are shown in Table \ref{tab:citation_contexts_exapmles}. This analysis supports the generally positive reception of the original study in the scholarly community.\\

\begin{table}[h!]
\caption{Examples of citation contexts categorized by class.}
\centering
\begin{tabular}{|c|p{10cm}|}
\hline
\textbf{Class} & \textbf{Citation Context Example} \\
\hline
Positive & ``Drawing inspiration from this study, the structure of animal visual systems mentioned above, and recent related work \textbf{(Ye et al. 2020)}, we derive two invariance properties that enhance the training of deep neural networks." \\
\hline
Neutral & ``Some complementary information can be found in recent works \textbf{(Ye et al. 2020)}" \\
\hline
\end{tabular}
\label{tab:citation_contexts_exapmles}
\end{table}

\section{Scope of reproducibility}
\label{sec:claims}
Our work seeks to evaluate and verify the primary claim of the original paper, namely that when network deconvolution is used instead of batch normalization in a deep learning architecture, the performance is generally improved \cite{ye2020network}. Verification of the claim required repeating the original experiments and analyses using the same methodologies, datasets, and computational tools as described in the original study. %Our objective is to determine whether the results reported in \cite{ye2020network} can be reproduced under the same conditions. 
The motivation for this reproducibility study extends beyond simply reproducing the original results. Our work also examines the consistency of the reported values in Table 1 of the original study by re-running the experiment multiple times under the same configurations. This approach is necessary because the performance of deep learning models, despite their deterministic nature, may vary due to random factors such as stochastic gradient descent and data shuffling. By performing repeated experiments, we can assess whether the reported improvements are consistently achievable or if they might be influenced by favorable random initializations. We document discrepancies and challenges encountered during the process, providing a comprehensive account of our findings.\\

\section{Methodology}

\rr{

We use the code provided by the original study \cite{githubGitHubYechengxideconvolution}. The code base requires Python GPU-based processing capabilities and the PyTorch framework. We discuss these requirements in detail in Computational Requirements, Subsection \ref{sec:Computational-requirements}. In Table 1 of the original study, authors reported 12 values, specifically 6 for BN and 6 for network deconvolution, for each CNN architecture under three different epoch settings. For a single CNN architecture, we conducted three attempts for each value reported in the original study (using the same hyperparameter setup), resulting in 36 reproduced values, to examine the model's consistency. Therefore we obtained 360 values for 10 architectures (Table \ref{tab:our_approach_rep_table_1}). Additionally, we reproduced the 14 values from the original study Table 2.}

\begin{table}[h!]
\caption{Our approach in reproducing the results from the original study \cite{ye2020network}: Table 1 (BN: Batch Normalization, ND: Network Deconvolution, rep.: reproduced/reproducing )}
\centering
% \begin{tabular}{|c|p{10cm}|}
\begin{tabular}{ l c c c c c}
\hline

\multirow{2}{*}{Dataset} & \multirow{2}{*}{Technique} & \multirow{2}{*}{\# tested} & \multirow{2}{*}{\# epoch settings}  & \multirow{2}{*}{\# rep. attempts}  & \multirow{2}{*}{Total rep.}   \\ 
 \multirow{2}{*}{} & \multirow{2}{*}{} & \multirow{2}{*}{CNNs} & \multirow{2}{*}{} & \multirow{2}{*}{per reported value} & \multirow{2}{*}{values} \\\\
 % \hline
 % \vspace{20mm}
 \hline 
 \multirow{2}{*}{CIFAR-10} & BN & 10 & 3 & 3 & 90\vspace{1mm}  \\ 
 \multirow{2}{*}{} & ND & 10 & 3 & 3 & 90\vspace{1mm}  \\
 
 \hline
 \multirow{2}{*}{CIFAR-100} & BN & 10 & 3 & 3 & 90\vspace{1mm} \\ 
 \multirow{2}{*}{} & ND & 10 & 3 & 3 & 90\vspace{1mm}  \\
 \hline
\end{tabular}
\label{tab:our_approach_rep_table_1}
\end{table}

\subsection{Datasets}
The authors of the original study have tested the proposed model and compared it against the state-of-the-art against three benchmark datasets:

\begin{enumerate}
    \item CIFAR-10: A dataset consisting of 60,000 32x32 color images across 10 classes. It is split into 50,000 training images and 10,000 test images (Size: 163 MB) \cite{cifar10}.
    \item CIFAR-100: A dataset with 60,000 32x32 color images spanning 100 classes. Similar to CIFAR-10, this dataset is divided into 50,000 training images and 10,000 test images (500 training and 100 testing images for each class) (Size: 161 MB) \cite{cifar100}.    
    \item ImageNet ILSVRC: \kt{The dataset released for ImageNet Large Scale Visual Recognition Challenge (ILSVRC)\footnote{\url{https://www.image-net.org/challenges/LSVRC/2012/index.php}}  2012 with 1.2 million training images, 50,000 validation images, and 100,000 test images. This dataset contains 1000 object classes. (Size: ~167.62 GB)} \cite{deng2009imagenet}.
\end{enumerate}

The results of CIFAR-10 and CIFAR-100 are shown in Table 1 in the original study. The results of ImageNet are shown in Table 2 in the original study. 

\subsection{Hyperparameters}
We used the same hyperparameter settings as those employed in the original study (see our study Table \ref{tab:Hyperparameter-settings}) :

\begin{table}[htbp]
    \centering
    \caption{Hyperparameter settings used to reproduce results from the original study.}
    \begin{tabular}{cccccc}
    \toprule
         \textbf{Original Study} & \textbf{Learning rate}  & \textbf{Optimizer} & \textbf{Batch size} & \textbf{Stride} & \textbf{Epochs} \\
    \midrule
        \textbf{Table 1} & 0.1 & SGD & 128 & 3 & [1,20,100]\\
        \textbf{Table 2} & [ 0.01, 0.1, 0.1 ] & SGD & 256 & 3 & 90 \\
    \bottomrule
    \end{tabular}
    \label{tab:Hyperparameter-settings}
\end{table}

% \subsubsection{Table 1}

% \begin{itemize}
%     \item Learning rate (start): 0.1
%     \item Optimizer: Stochastic Gradient Descent (SGD)
%     \item Batch size: 128
%     \item Stride: 3
%     \item Epochs : [1, 20,100]
% \end{itemize}

% \subsubsection{Table 2}
% \rr{\begin{itemize}
%     \item Learning rate (start): \{ VGG-11: 0.01, ResNet-18 \& DenseNet-121 :0.1 \}
%     \item Optimizer: Stochastic Gradient Descent (SGD)
%     \item Batch size: 256
%     \item Stride: 3
%     \item Epochs : 90
% \end{itemize}}

\subsection{Preparing the ImageNet dataset}\label{Preparing-the-ImageNet-dataset}

The CIFAR-10 and CIFAR-100 datasets required for reproducing Table 1 of the original study are automatically downloaded through the codebase using torchvision \cite{pytorchTorchvisiondatasetsx2014}, eliminating the need for manual dataset preparation. The ImageNet ILSVRG \cite{deng2009imagenet} subset was required to reproduce the results reported in Table 2 of the original study. Downloading the ImageNet ILSVRC dataset is no longer possible from the original source on the ImageNet website\footnote{\url{https://www.image-net.org/}}. The URL redirected to a new download link hosted on Kaggle.\footnote{\label{kaggle_}\url{https://www.kaggle.com/c/imagenet-object-localization-challenge/data}}. After inspection, we found that the folder structure of the Kaggle version is different from the original data used by the original paper. Therefore, we followed the below steps to modify the folder structure to make it compatible with the input format required by the code (the detail is described at our GitHub repository \url{https://github.com/lamps-lab/rep-network-deconvolution}).

\begin{enumerate}
    
    \item Clone the GitHub repository  \url{https://github.com/lamps-lab/rep-network-deconvolution} to a workspace on a local server (hereafter \textbf{\textit{workspace}}). This \textbf{\textit{workspace}} is an extended fork of the GitHub repository\footnote{\url{https://github.com/yechengxi/deconvolution}} provided by the authors of the original study.
    \item Download the ``\textit{ILSVRC.zip}" file from Kaggle Imagenet-Object-Localization-Challenge \url{https://www.kaggle.com/c/imagenet-object-localization-challenge/data}.
    
    \item Inside the `\textit{imagenet}' folder of the \textbf{\textit{workspace}}, unzip the ``\textit{ILSVRC.zip}".
    \item Move the ``\textit{valprep.sh}” inside ``\textit{imagenet/ILSVRC/Data/CLS-LOC/val}” folder which is initially available in the root directory of the \textbf{\textit{workspace}} and execute ``\textit{valprep.sh}” (on successful execution, you should see the created 1000 sub-directories inside the same `val' folder representing the 1000  object categories as described in our GitHub repository\footnote{\url{https://github.com/lamps-lab/rep-network-deconvolution}}).
    
\end{enumerate}

\subsection{Experimental setup}

After preparing the datasets, we follow the steps below to set up the experimental environment. 

\begin{enumerate}
    % \item Clone the GitHub repository provided by the authors: \url{https://github.com/yechengxi/deconvolution}
    \item Create a python virtual environment following the instructions from \url{https://docs.python.org/3/library/venv.html} and activate it.
    \item Install the python dependencies using the ``\textit{requirements.txt}'' file (available in the root directory of the \textbf{\textit{workspace}}).  
    \item Perform initial testing using the ``\textit{deconv\_rep.ipynb}'' notebook, which is also available in the root directory of the \textbf{\textit{workspace}}, with 1 epoch to verify the code's functionality (GPU processing capabilities required).
    \item Conduct more extensive experiments using the bash scripts under the \textbf{\textit{workspace}} on a local GPU cluster.
\end{enumerate}

\subsection{Computational requirements}
\label{sec:Computational-requirements}
% We incorporated the following hardware and software:\\

Table 1 in the original paper was reproduced on an HPC cluster equipped with an NVIDIA V100 GPU with 16GB memory. Because ImageNet is significantly larger than CIFAR-10 and CIFAR-100, we moved the experiments to a more powerful server with NVIDIA A100, providing 80GB GPU. The original study does not specify the software versions used, and several software packages have been upgraded since the paper was published. Below is the list of resources we utilized in our study.\\

\textbf{Hardware:}

\begin{itemize}
    \item Local GPU Cluster: 
        \begin{itemize}[label=$\diamond$]
            \item 16 GB GPU – NVIDIA V100 (For original study Table 1)
            \item 80 GB GPU – NVIDIA A100 (For original study Table 2)
        \end{itemize}
    \item Disk : \rr{2 TB (Shared for both Table 1 and 2 of the original study)} 
    \item CPU : \rr{Intel(R) Xeon(R) Gold 6148 CPU @ 2.40GHz (multi-core)}
    \item RAM: \rr{64 GB (Shared for both Table 1 and 2 of the original study)} 
\end{itemize}

\textbf{Software:}
\begin{itemize}
    \item \rr{Operating System: Rocky Linux 9.3 (Blue Onyx)}
    \item Python 3.10.2
    \item SciPy 1.10.1
    \item NumPy 1.23.5
    \item TensorBoard 2.12.0
    \item Matplotlib 3.7.1
    \item PyTorch 1.13
    \item Torchvision 0.14.1
    \item TensorFlow 2.12.0
    \item Slurm (for job scheduling on the internal GPU cluster)
\end{itemize}

\subsection{Recommendations for Minimal Computational Requirements}
\label{sec:Recommendations-for-Minimal-Computational-Requirements}
\rr{
For researchers attempting to reproduce the original study, we recommend the following minimum hardware requirements:

\begin{itemize}
  \item GPU: 16 GB GPU memory for Table 1, 40 GB GPU memory for Table 2
  \item CPU: Multi-core processor, such as Intel Xeon or AMD Ryzen
  \item RAM: 16 GB
  \item Disk: At least 240 GB for data and model storage
\end{itemize}

}

\section{Results}

\subsection{Reproduced Results of Table 1 in the Original paper}

\begin{figure}[H] %[htbp]
  \centering
    \setlength{\fboxsep}{2pt} % Adjust the padding
    \setlength{\fboxrule}{0.2pt} % Adjust the border width
    \fbox{\includegraphics[width=1\linewidth]{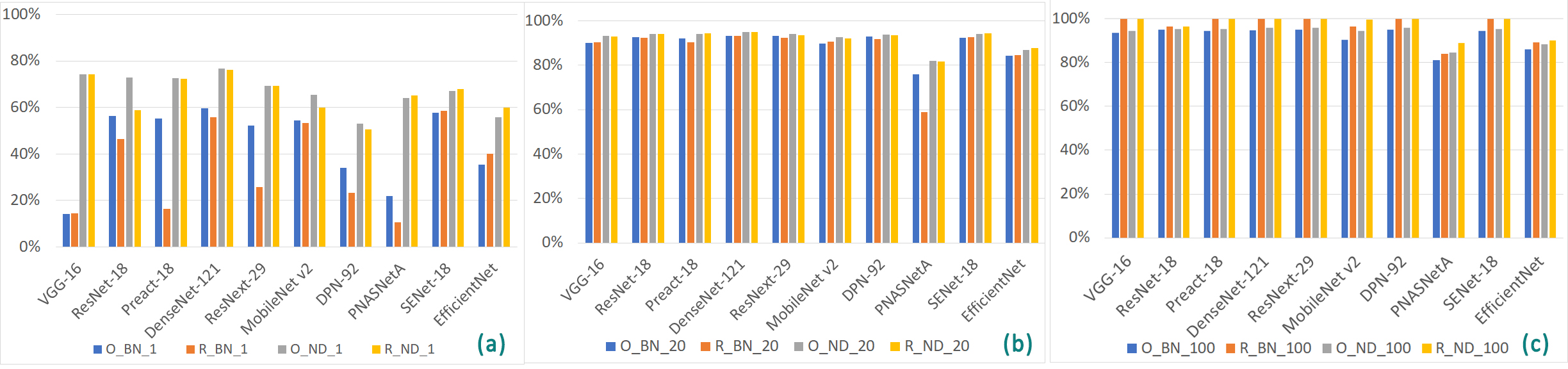}}
    % Adjust width as needed
    \caption{\textbf{Comparing original and reproduced accuracy for {\normalsize{CIFAR-10}} dataset: Original Study \cite{ye2020network} - Table 1 (a) with 1 epoch, (b) with 20 epochs, (c) with 100 epochs. (Legend: O\_BN\_20 - original accuracy using batch normalization with 20 epochs, R\_ND\_100 - reproduced accuracy using network deconvolution with 100 epochs).}}
    \label{fig:result-figures-cifar10}
\end{figure}

\begin{figure}[H] %[htbp]

  \centering
    \setlength{\fboxsep}{2pt} % Adjust the padding
    \setlength{\fboxrule}{0.2pt} % Adjust the border width
    \fbox{\includegraphics[width=1\linewidth]{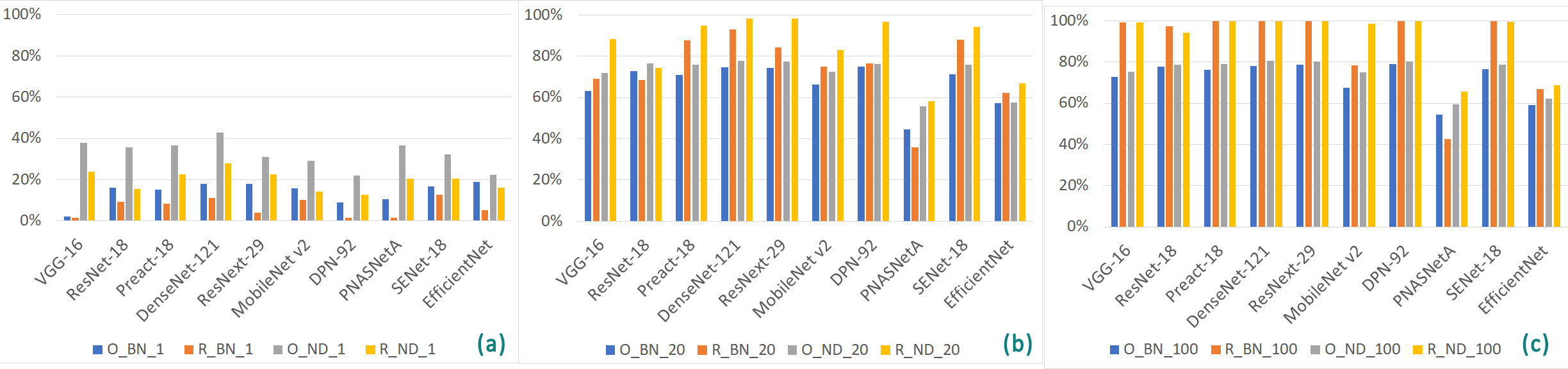}}
    % Adjust width as needed
    \caption{\textbf{Comparing original and reproduced accuracy for {\normalsize{CIFAR-100}} dataset: Original Study \cite{ye2020network} - Table 1 (a) with 1 epoch, (b) with 20 epochs, (c) with 100 epochs. Labels: (Legend: O\_BN\_20 - original accuracy using batch normalization with 20 epochs, R\_ND\_100 - reproduced accuracy using network deconvolution with 100 epochs).}}
    \label{fig:result-figures-cifar100}
\end{figure}

The authors of the original paper reported the model performance using ``Accuracy'' as the metric. We report the reproduced accuracy for each architecture using each of the three datasets, namely, CIFAR10, CIFAR100, and ImageNet. To investigate the consistency, we obtained 3 results per one value reported in the original study Table 1. For example, we ran the experiment under the same hyper-parameter settings for batch normalization for 1 epoch for a particular architecture three times and recorded the results per each attempt. Then we averaged the results of three attempts and compared them with the original study's reported values. Table \ref{tab:results_cifar10} and Figure \ref{fig:result-figures-cifar10} compare the averaged reproduced results for CIFAR-10 dataset with the original study's values. A similar comparison was made for CIFAR-100 in Table \ref{tab:results_cifar100} and Figure \ref{fig:result-figures-cifar100}.

\begin{table}[H]
    \caption{\textbf{Results from the original study Table 1 and the reproduced averaged values from our study for {\normalsize{CIFAR-10}} dataset with 1, 20, and 100 epochs. Architectures: (1) VGG-16, (2) ResNet-18, (3) Preact-18, (4) DenseNet-121, (5) ResNext-29, (6) MobileNet v2, (7) DPN-92, (8) PNASNet-18, (9) SENet-18, (10) EfficientNet (All values are presented as percentages). Color codes: \R{Red} - If the reproduced result is lower than the original value by more than 10\%, \g{Green} - If the reproduced value is greater than the original value, Black - If the reproduced value is less than the original value, but the difference between two values is no more than 10\%. }}
    \centering

\begin{tabular}{p{0.05\linewidth}p{0.05\linewidth}p{0.05\linewidth}|p{0.05\linewidth}p{0.05\linewidth}|p{0.05\linewidth}
p{0.05\linewidth}|p{0.05\linewidth}p{0.05\linewidth}|p{0.05\linewidth}p{0.05\linewidth}|p{0.05\linewidth}p{0.05\linewidth}}
  \hline
  \multirow{3}{*}{\textbf{Arch.}} & \multicolumn{12}{c}{\textbf{CIFAR-10}} \\
  %\cline{2-13}
   \cmidrule{2-13}
  \multirow{3}{*}{}  & \multicolumn{2}{c}{BN 1} & \multicolumn{2}{c}{ND 1} & \multicolumn{2}{c}{BN 20} & \multicolumn{2}{c}{ND 20} & \multicolumn{2}{c}{BN 100} & \multicolumn{2}{c}{ND 100}\\
  %\cline{2-13}
   \cmidrule{2-13}
    \multirow{3}{*}{} & Org. Value & Rep. Avg & Org. Value & Rep. Avg & Org. Value & Rep. Avg & Org. Value & Rep. Avg & Org. Value & Rep. Avg & Org. Value & Rep. Avg \\
    \cmidrule{1-13}
    $(1)$ & 14.12 & \g{14.38} & 74.18 & \g{74.20} & 90.07 & \g{90.36} & 93.25 & 92.83 & 93.58 & \g{99.86} & 94.56 & \g{99.88} \\
    $(2)$ & 56.25 & 46.56 & 72.89 & \R{58.69} & 92.64 & 92.35 & 94.07 & \g{94.20} & 94.87 & \g{96.36} & 95.40 & \g{96.42} \\
    $(3)$ & 55.15 & \R{16.42} & 72.70 & 72.33 & 91.93 & 90.32 & 94.10 & \g{94.26} & 94.37 & \g{99.92} & 95.44 & \g{99.96} \\
    $(4)$ & 59.56 & 55.78 & 76.63 & 76.31 & 93.25 & 93.11 & 94.89 & 94.87 & 94.71 & \g{99.95} & 95.88 & \g{99.97} \\
    $(5)$ & 52.14 & \R{25.70} & 69.22 & \g{69.35} & 93.12 & 92.34 & 94.05 & 93.57 & 95.15 & \g{99.99} & 95.80 & \g{99.97} \\
    $(6)$ & 54.29 & 53.26 & 65.40 & 59.85 & 89.86 & \g{90.70} & 92.52 & 92.11 & 90.51 & \g{96.34} & 94.35 & \g{99.55} \\
    $(7)$ & 34.00 & \R{23.27} & 53.02 & 50.61 & 92.87 & 91.79 & 93.74 & 93.55 & 95.14 & \g{99.96} & 95.82 & \g{99.96} \\
    $(8)$ & 21.81 & \R{10.55} & 64.19 & \g{65.19} & 75.85 & \R{58.84} & 81.97 & 81.61 & 81.22 & \g{84.12} & 84.45 & \g{88.98} \\
    $(9)$ & 57.63 & \g{58.62} & 67.21 & \g{67.80} & 92.37 & \g{92.74} & 94.11 & \g{94.25} & 94.57 & \g{99.95} & 95.38 & \g{99.95} \\
    $(10)$ & 35.40 & \g{40.10} & 55.67 & \g{59.91} & 84.21 & \g{84.68} & 86.78 & \g{87.77} & 86.07 & \g{89.16} & 88.42 & \g{90.10} \\
   \hline
\end{tabular}
\label{tab:results_cifar10}
\end{table}

\begin{table}[H]
    \caption{\textbf{Results from the original study Table 1 and the reproduced averaged values from our study for {\normalsize{CIFAR-100}} dataset with 1, 20, and 100 epochs. Architectures: (1) VGG-16, (2) ResNet-18, (3) Preact-18, (4) DenseNet-121, (5) ResNext-29, (6) MobileNet v2, (7) DPN-92, (8) PNASNet-18, (9) SENet-18, (10) EfficientNet (all values are presented as percentages).  Color codes: Same as Table \ref{tab:results_cifar10}}}
    \centering

\begin{tabular}{p{0.05\linewidth}p{0.05\linewidth}p{0.05\linewidth}|p{0.05\linewidth}p{0.05\linewidth}|p{0.05\linewidth}
p{0.05\linewidth}|p{0.05\linewidth}p{0.05\linewidth}|p{0.05\linewidth}p{0.05\linewidth}|p{0.05\linewidth}p{0.05\linewidth}}
  \hline
  \multirow{3}{*}{\textbf{Arch.}} & \multicolumn{12}{c}{\textbf{CIFAR-100}} \\
  %\cline{2-13}
   \cmidrule{2-13}
  \multirow{3}{*}{}  & \multicolumn{2}{c}{BN 1} & \multicolumn{2}{c}{ND 1} & \multicolumn{2}{c}{BN 20} & \multicolumn{2}{c}{ND 20} & \multicolumn{2}{c}{BN 100} & \multicolumn{2}{c}{ND 100}\\
  %\cline{2-13}
   \cmidrule{2-13}
    \multirow{3}{*}{} & Org. Value & Rep. Avg & Org. Value & Rep. Avg & Org. Value & Rep. Avg & Org. Value & Rep. Avg & Org. Value & Rep. Avg & Org. Value & Rep. Avg \\
    \cmidrule{1-13}
    $(1)$ & 2.01 & 1.47 & 37.94 & \R{24.02} & 63.22 & \g{69.04} & 71.97 & \g{88.43} & 72.75 & \g{99.23} & 75.32 & \g{99.30} \\
    $(2)$ & 16.10 & 9.15 & 35.73 & \R{15.41} & 72.67 & 68.49 & 76.55 & 74.32 & 77.70 & \g{97.42} & 78.63 & \g{94.31} \\
    $(3)$ & 15.17 & 8.39 & 36.52 & \R{22.60} & 70.79 & \g{87.73} & 76.04 & \g{94.71} & 76.14 & \g{99.90} & 79.14 & \g{99.75} \\
    $(4)$ & 17.90 & 11.12 & 42.91 & \R{28.06} & 74.79 & \g{92.95} & 77.63 & \g{98.12} & 77.99 & \g{99.89} & 80.69 & \g{99.93} \\
    $(5)$ & 17.98 & \R{3.89} & 30.93 & \R{22.68} & 74.26 & \g{84.44} & 77.35 & \g{98.32} & 78.60 & \g{99.93} & 80.34 & \g{99.96} \\
    $(6)$ & 15.88 & 10.11 & 29.01 & \R{14.32} & 66.31 & \g{75.05} & 72.33 & \g{82.92} & 67.52 & \g{78.53} & 74.90 & \g{98.52} \\
    $(7)$ & 8.84 & 1.58 & 21.89 & 12.82 & 74.87 & \g{76.59} & 76.12 & \g{96.77} & 78.87 & \g{99.93} & 80.38 & \g{99.97} \\
    $(8)$ & 10.49 & 1.45 & 36.52 & \R{20.47} & 44.60 & 35.67 & 55.65 & \g{58.27} & 54.52 & \R{42.50} & 59.44 & \g{65.66} \\
    $(9)$ & 16.60 & 12.58 & 32.22 & \R{20.50} & 71.10 & \g{87.87} & 75.79 & \g{94.12} & 76.41 & \g{99.92} & 78.63 & \g{99.66} \\
    $(10)$ & 19.03 & \R{5.27} & 22.40 & 16.11 & 57.23 & \g{62.14} & 57.59 & \g{66.83} & 59.09 & \g{66.78} & 62.37 & \g{68.79} \\
   \hline
\end{tabular}
\label{tab:results_cifar100}
\end{table}

\vspace{-40pt}

The results show that the model performance with network deconvolution is always better than the model performance with batch normalization. To determine if the values are reproducible, we consider a 10\% reproducibility threshold for the accuracy metric as our criterion. In both CIFAR-10 and CIFAR-100 datasets 36 out of 60 values (60\% ) were reproduced with better results. In the ImageNet dataset 9 out of 14 values were reproduced with better results. There are few instances in CIFAR10 and CIFAR-100 datasets where the reproduced value is lower than the originally reported value. Most of these cases occur when the models were trained for 1 epoch. However, for the cases where the models were trained for 20 epochs and 100 epochs, the reproduced values show improved results, showing high accuracy values similar to those reported in the original study. The architecture PNASNet-18 shows weak performance in both batch normalization and network deconvolution as its accuracy values are comparatively low. \\\\We calculated the averaged squared deviation from the original value for the three reproduced values obtained at three attempts . We used this calculated value to understand the consistency of the reproducing results. Consistency, in this context, means that the deviation of the reproduced values from the originally reported value should be minimal. Therefore, if the averaged squared deviation of reproduced values fall below 0.5 threshold we consider that the reproduced results are consistent. For CIFAR-10 dataset the averaged squared deviation lies between 0.00 to 0.23 and for CIFAR-100 dataset it lies between 0.00 and 0.11. Therefore the reproduced results are consistent.

\subsection{Reproduced Results of Table 2 in the Original Paper}

\begin{table}[H]
    \caption{\textbf{Accuracy values reported by the original study Table 2 and the reproduced values for VGG-11 with 90 epochs (Rep.:Reproduced value).Color codes: \R{Red} - If the reproduced result is lower than the original value by more than 10\%, \g{Green} - If the reproduced value is greater than the original value, Black - If the reproduced value is less than the original value, but the difference between two values is no more than 10\%.}}
    \centering

\begin{tabular}{p{0.18\linewidth}p{0.08\linewidth}p{0.095\linewidth}p{0.08\linewidth}p{0.095\linewidth}p{0.08\linewidth}
p{0.095\linewidth}}
  \hline
  % \multirow{3}{*}{} & \multicolumn{6}{c}{\textbf{ImageNet}} \\
  % \cmidrule{2-7}
  \multirow{3}{*}{} & \multicolumn{6}{c}{\textbf{VGG-11}} \\
  \cmidrule{2-7}
  \multirow{3}{*}{} & Original & Original Rep. & BN & BN\hspace{0.7mm} Rep. & Deconv & Deconv\hspace{0.5mm} Rep. \\
  \cmidrule{1-7}
   ImageNet top 1 & 69.02 & \centering \g{69.61} & 70.38 & \g{71.35} & 71.95 & 71.95  \\
   ImageNet top 5 & 88.63 & \centering 89.08 & 89.81 & \g{90.25} & 90.49 & \g{90.60}  \\
   \hline
\end{tabular}
\label{tab:imagenet-vgg11}
\end{table}

\begin{table}[H]
    \caption{\textbf{Accuracy values reported by the original study's Table 2 and the reproduced values for the architectures ResNet-18 and DenseNet-121 with 90 epochs (Rep.:Reproduced value). Color codes: Same as Table \ref{tab:imagenet-vgg11}.}}
    \centering

\begin{tabular}{p{0.18\linewidth}p{0.06\linewidth}p{0.06\linewidth}p{0.08\linewidth}p{0.08\linewidth}|p{0.06\linewidth}p{0.06\linewidth}p{0.08\linewidth}p{0.08\linewidth}}
  \hline
  % \multirow{3}{*}{} & \multicolumn{8}{c}{\textbf{ImageNet}} \\
  % \cmidrule{2-9}
  \multirow{3}{*}{} & \multicolumn{4}{c}{\textbf{ResNet-18}} & \multicolumn{4}{c}{\textbf{DenseNet-121}} \\
  \cmidrule{2-9}
  \multirow{3}{*}{} & BN & BN Rep. & Deconv & Deconv Rep. & BN & BN Rep. & Deconv & Deconv Rep. \\
  \cmidrule{1-9}
   ImageNet top 1 & 69.76 & \g{70.72} & 71.24 & \g{71.58} & 74.65 & \g{75.88} & 75.73 & 75.67 \\
   ImageNet top 5 & 89.08 & \g{89.73} & 90.14 & 90.27 & 92.17 & \g{92.96} & 92.75 & 92.69  \\
   \hline
\end{tabular}
\label{tab:imagenet-resnet-densenet}
\end{table}

We reported the top-1 and top-5 accuracy values for BN and network deconvolution. Reproduced results for the VGG-11, ResNet-18, and DenseNet-121 using the ImageNet dataset are consistent since they fall within our reproducibility threshold. Therefore it confirms the main claim in the original study (Table \ref{tab:results_cifar10}, \ref{tab:results_cifar100}, \ref{tab:imagenet-vgg11}, and \ref{tab:imagenet-resnet-densenet}). 

% \begin{table}[H]
%     \caption{\textbf{Results from the original study Table 02 and the reproduced values for architecture DenseNet-121}}
%     \centering

% \begin{tabular}{p{0.15\linewidth}p{0.08\linewidth}p{0.08\linewidth}p{0.08\linewidth}p{0.08\linewidth}}
%   \hline
%   \multirow{3}{*}{} & \multicolumn{4}{c}{\textbf{ImageNet}} \\
%   \cmidrule{2-5}
%   \multirow{3}{*}{} & \multicolumn{4}{c}{DenseNet-121} \\
%   \cmidrule{2-5}
%   \multirow{3}{*}{} & BN & BN. rep & ND & ND. rep \\
%   \cmidrule{1-5}
%    ImageNet top 1 & 74.65 & 75.88 & 75.73 & 75.67 \\
%    ImageNet top 5 & 92.17 & 92.96 & 92.75 & 92.69 \\
%    \hline
% \end{tabular}
% \label{tab:results_2}
% \end{table}

\subsection{Training Time Comparison}
\begin{figure}[H] %[htbp]
  \centering
    \setlength{\fboxsep}{2pt} % Adjust the padding
    \setlength{\fboxrule}{0.2pt} % Adjust the border width
    \fbox{\includegraphics[width=1\linewidth]{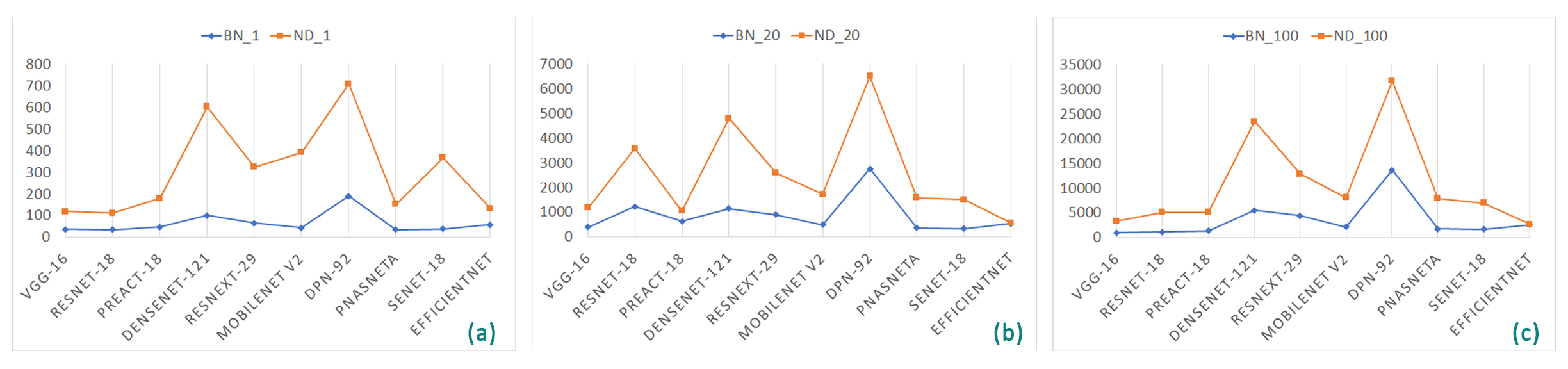}}
    % Adjust width as needed
    \caption{\textbf{Training times for each CNN architecture with CIFAR-10 dataset: (a) with 1 epoch, (b) with 20 epochs, (c) with 100 epochs}}
    \label{fig:result-figures-time-cifar10}
\end{figure}

\begin{figure}[H] %[htbp]
  \centering
    \setlength{\fboxsep}{2pt} % Adjust the padding
    \setlength{\fboxrule}{0.2pt} % Adjust the border width
    \fbox{\includegraphics[width=1\linewidth]{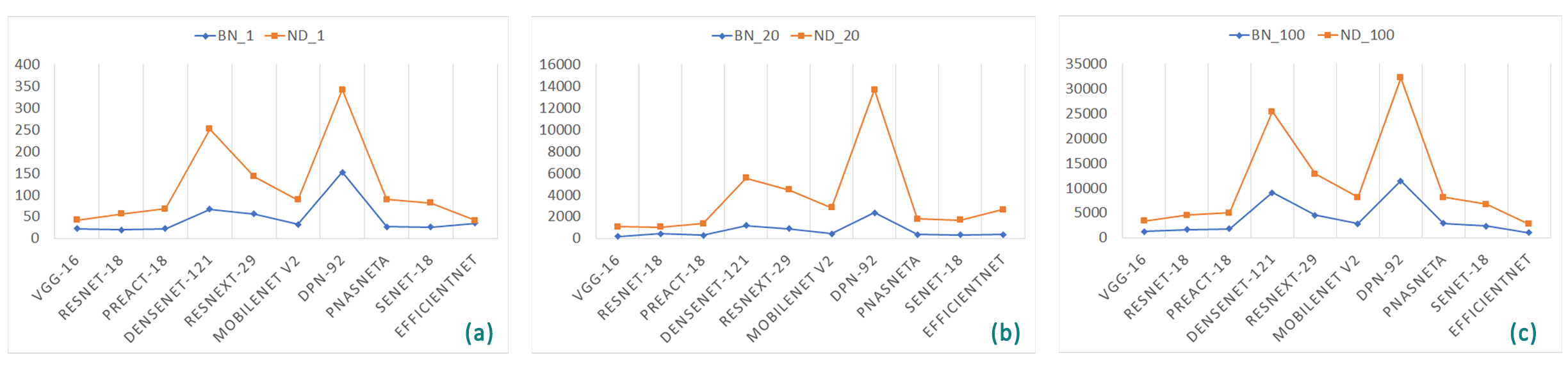}}
    % Adjust width as needed
    \caption{\textbf{Training times for each CNN architecture with CIFAR-100 dataset: (a) with 1 epoch, (b) with 20 epochs, (c) with 100 epochs}}
    \label{fig:result-figures-time-cifar100}
\end{figure}

During our reproducibility study, we observed a remarkable difference in the training time between  BN and network deconvolution, which was not reported in the original paper. Training time is often a factor to consider when building a deep learning architecture. In this section, we compare the training times of the BN and network convolution observed when testing them on 10 deep learning architectures. 
The results of CIFAR-10 are shown in Table \ref{tab:cifar10-train-times}  and Figure \ref{fig:result-figures-time-cifar10}. The results of CIFAR-100 are shown in Table \ref{tab:cifar100-train-times} and Figure \ref{fig:result-figures-time-cifar100}.  (Available at our GitHub repository \footnote{\url{https://github.com/lamps-lab/rep-network-deconvolution}} ``\textit{results/rep\_values.xlsx}").\\

\begin{table}[H]
    \caption{\textbf{Training time reported for the batch normalization and network deconvolution for CIFAR-10 dataset with 1, 20, 100 epochs. (Measuring unit: seconds).}}
    \centering
    \begin{tabular}{p{0.18\linewidth}p{0.08\linewidth}p{0.08\linewidth}|p{0.08\linewidth}p{0.08\linewidth}|p{0.08\linewidth}p{0.1\linewidth}}
  \hline
  \multirow{3}{*}{\textbf{Architecture}} & \multicolumn{6}{c}{\textbf{CIFAR-10}} \\
  %\cline{2-13}
   \cmidrule{2-7}
  \multirow{3}{*}{}  & BN 1 & ND 1 & BN 20 & ND 20 & BN 100 & ND 100\\
    \cmidrule{1-7}
    VGG-16 & 37.38 & 118.69 & 407.35 & 1183.07 & 990.47 & 3356.51 \\
    ResNet-18 & 35.41 & 112.63 & 1235.03 & 3586.99 & 1122.20 & 5142.16 \\
    Preact-18 & 47.88 & 180.79 & 639.52 & 1061.18 & 1356.69 & 5119.86 \\
    DenseNet-121 & 101.96 & 603.26 & 1156.61 & 4792.65 & 5549.08 & 23524.84 \\
    ResNext-29 & 66.39 & 324.62 & 906.48 & 2595.79 & 4473.94 & 12885.14 \\
    MobileNet v2 & 43.88 & 393.73 & 492.24 & 1738.16 & 2183.71 & 8106.39 \\
    DPN-92 & 190.40 & 708.68 & 2768.02 & 6500.02 & 13646.93 & 31776.99 \\
    PNASNet-18 & 35.22 & 153.21 & 369.79 & 1593.17 & 1811.29 & 7948.39 \\
    SENet-18 & 39.52 & 367.79 & 344.22 & 1517.79 & 1670.13 & 7020.19 \\
    EfficientNet & 58.05 & 133.67 & 550.25 & 578.58 & 2647.03 & 2712.74 \\
   \hline
\end{tabular}
\label{tab:cifar10-train-times}
\end{table}

\begin{table}[H]
    \caption{\textbf{Training time reported for the batch normalization and network deconvolution for CIFAR-100 dataset with 1, 20, 100 epochs. (Measuring unit: seconds).}}
    \centering
    \begin{tabular}{p{0.18\linewidth}p{0.08\linewidth}p{0.08\linewidth}|p{0.08\linewidth}p{0.08\linewidth}|p{0.08\linewidth}p{0.1\linewidth}}
  \hline
  \multirow{3}{*}{\textbf{Archiecture}} & \multicolumn{6}{c}{\textbf{CIFAR-100}} \\
  %\cline{2-13}
   \cmidrule{2-7}
  \multirow{3}{*}{}  & BN 1 & ND 1 & BN 20 & ND 20 & BN 100 & ND 100\\
    \cmidrule{1-7}
    VGG-16 & 22.54 & 42.85 & 204.81 & 1104.37 & 1274.84 & 3424.08 \\
    ResNet-18 & 20.17 & 56.87 & 446.33 & 1076.18 & 1713 & 4589.303 \\
    Preact-18 & 22.70 & 68.50 & 294.96 & 1370.68 & 1856.72 & 5098.77 \\
    DenseNet-121 & 68.12 & 252.16 & 1193.88 & 5578.77 & 9109.30 & 25429.58 \\
    ResNext-29 & 56.71 & 142.49 & 909.63 & 4498.96 & 4574.03 & 12895.70 \\
    MobileNet v2 & 32.96 & 89.18 & 449.72 & 2845.72 & 2880.88 & 8181.95 \\
    DPN-92 & 158.08 & 342.03 & 2385.98 & 13684.59 & 11472.35 & 32227.37 \\
    PNASNet-18 & 27.09 & 89.98 & 378.45 & 1815.56 & 2956.37 & 8219.82 \\
    SENet-18 & 26.56 & 81.89 & 347.26 & 1691.54 & 2433.31 & 6813.16 \\
    EfficientNet & 35.5 & 41.53 & 387.08 & 2683.59 & 1090.063 & 2784.08 \\
   \hline
\end{tabular}
\label{tab:cifar100-train-times}
\end{table}

% \subsubsection{Training times for each CNN architecture with CIFAR-10 dataset}

% \kt{update the graphs}

% \begin{center}
% \begin{minipage}{0.3\textwidth}
% \includegraphics[width=\linewidth]{figures/cifar10_1.png}
%     % \caption{}{\small Model train times (1 epoch) }
% \end{minipage}\hfill
% \begin{minipage}{0.3\textwidth}
% \includegraphics[width=\linewidth]{figures/cifar10_20.png} 
% % \captionof{}{\small Model train times (20 epochs) }   
% \end{minipage}\hfill
% \begin{minipage}{0.3\textwidth}
% \includegraphics[width=\linewidth]{figures/cifar10_100.png}
% % \captionof{}{\small Model train times (100 epochs) }
% \end{minipage}
% \label{name_label}
% \end{center}

% \subsubsection{Training times for each CNN architecture with CIFAR-100 dataset}

% \begin{center}
% \begin{minipage}{0.3\textwidth}
% \includegraphics[width=\linewidth]{figures/cifar100_1.png}
% % \captionof{}{\small Model train times (1 epoch)}
% \end{minipage}\hfill
% \begin{minipage}{0.3\textwidth}
% \includegraphics[width=\linewidth]{figures/cifar100_20.png} 
% % \captionof{}{\small Model train times (20 epochs)}   
% \end{minipage}\hfill
% \begin{minipage}{0.3\textwidth}
% \includegraphics[width=\linewidth]{figures/cifar100_100.png}
% % \captionof{}{\small Model train times (100 epochs)}
% \end{minipage}
% \label{name_label}
% \end{center}

Figures \ref{fig:result-figures-time-cifar10} and \ref{fig:result-figures-time-cifar100} depict that the training time of architectures with network deconvolution is always higher than the architectures with BN. Remarkable time gaps are visible in DenseNet-121 and DPN-92. DPN-92 always exhibits the longest training time for both BN and network deconvolution and the biggest difference between them. The shortest time is seen in the EfficientNet for the CIFAR-10. The same trend was observed in the CIFAR-100 dataset except ResNet-18 has the shortest time difference in 20 epochs.

% Further, reproduced results confirmed the original study's claim, as we observed improved performance when using Network Deconvolution instead of Batch Normalization across the various architectures and datasets evaluated. While there were some minor deviations in accuracy (within 10\%) for Table 01 of the original study, the overall trend was consistent with the original paper's findings.

% \subsubsection{Training times for each CNN architecture with ImageNet dataset (Table 02)}
% \kt{add a table with training times of imagenet dataset}

\begin{table}[htbp]
    \caption{\textbf{Model training times for the ImageNet dataset. (Measurements explanation: d-day, h-hour, min-minutes) }}
    \centering
    \begin{tabular}{p{0.11\linewidth}p{0.11\linewidth}p{0.11\linewidth}|p{0.11\linewidth}p{0.11\linewidth}|p{0.11\linewidth}p{0.11\linewidth}}
      % \hline
      % \multicolumn{7}{c}{\textbf{ImageNet}} \\
      \cmidrule{1-7}
      \multicolumn{3}{c}{\textbf{VGG-11}} & \multicolumn{2}{c}{\textbf{ResNet-18}} & \multicolumn{2}{c}{\textbf{DenseNet-121}} \\
      \cmidrule{1-7}
      Original Rep. & BN Rep. & ND Rep. & BN Rep. & ND Rep. & BN. rep & ND Rep. \\
      \cmidrule{1-7}
       1d,7h, 23min & 1d,15h, 38min & 2d,21h, 27min & 0d,13h, 8 min & 1d,11h, 48 min & 3d,21h, 18min & 5d,14h, 10min  \\
        \hline
    \end{tabular}
\label{tab:imagenet-train-times}
\end{table}

Table \ref{tab:imagenet-train-times} shows the training times for the ImageNet dataset using the VGG-11, ResNet-18, and DenseNet-121 architectures. In this table, we report the time taken to reproduce each value stated in our study, as presented in Tables \ref{tab:imagenet-vgg11} and \ref{tab:imagenet-resnet-densenet}. 

% Additionally, we report the training time of the original network for VGG-11 architecture (Original Rep.) following the original study Table 2 (Our study Table 6).

% \subsection{Results summary}
% \subsubsection{Original Paper Table 01}
% \begin{itemize}
%     \item When comparing results of batch normalization with network deconvolution, performance always show improvement with network deconvolution.
%     \item Overall, reproduced results are consistent for epochs 20 and 100.
%     \item In epoch 20 and epoch 100 reproduced values show improvement in each model architecture.
%     \item With only 1 epoch, most of the reproduced results are lower than the originally reported values.
%     \item When using only 20 epochs with network deconvolution, the results are similar to those obtained using 100 epochs with batch normalization.
%     \item The time taken to train the models are high for network deconvolution in all architectures
%     \item The architecture PNASNetA  show fairly low accuracy values in both CIFAR-10 and CIFAR-100 datasets.
%     %Optimizer SGD and learning rate 0.1 may not be the best hyper parameters for this architecture
% \end{itemize}

% \kt{\subsubsection{Original Paper Table 02}
% \begin{itemize}
%     \item The reproduced results of batch normalization and network deconvolution with the ImageNet dataset are almost the same or slightly increased when compared to the original values. 
%     \item The performance of the models increased with network deconvolution than with batch normalization in all three architectures.  
% \end{itemize}
% }

\section{Discussion}

In this work, we rigorously investigated the reproducibility of the reported results of the original paper by conducting experiments under the same hyperparameter settings. Specially, for Table 1 of the original study, we ran each architecture three times under the same hyperparameter configurations. The results of each run slightly vary due to the random initialization but majority of the results fall within the 10\% reproducibility threshold. By systematically examining the model's performance under varying conditions, we aimed to provide insights on reprocubility and consistency of the reproduced results. \\\\
The reproduced accuracies for Table 1  and Table 2 of the original paper are nearly the same or even higher when compared to the original values. Among 134 reported results , 116 results (\g{green} or black) fall within the threshold of 10\% (relative error) and 80 results (\g{green}) are better than the original values. This improvement is systematic and does not change the conclusion. 
% All experimental results verify the claim of the original study that network deconvolution outperforms batch normalization to enhance the performance of various model architectures. \\
In our investigation, we identified several factors that could potentially cause the systematic improvement of observed results. Notably, the advancement in numerical stability of the libraries and frameworks, such as the upgrades from NumPy 1.16.1 to NumPy 1.23.5 and from PyTorch 1.0 to PyTorch 1.13, may have contributed to the systematic improvements. Additionally, the improvement can also be caused by the implementation of optimization algorithms and parallelism in TensorFlow. \\\\
The calculated averaged squared deviation from the original values for each architecture during the three attempts show that all the results are consistent. The Training time comparison shows that network deconvolution in general requires more training time compared to BN but the performance improvements outweigh the increased training time. \\\\
Network deconvolution has been recognized and adopted in many in contemporary architectures, including U-Net\cite{unet-ronneberger2015unetconvolutionalnetworksbiomedical}, SegNet\cite{segnet}, GANs\cite{gans-goodfellow2014generativeadversarialnetworks}, Super-resolution networks\cite{srn-guo2023asconvsrfastlightweightsuperresolution}, Pix2Pix\cite{pix2pix-isola2018imagetoimagetranslationconditionaladversarial}, and DeepLab\cite{deeplab}. In most cases, this technique is used to up-sample the convoluted images to produce high-quality images.

\subsection{What was easy}
% Give your judgement of what was easy to reproduce. Perhaps the author's code is clearly written and easy to run, so it was easy to verify the majority of original claims. Or, the explanation in the paper was really easy to follow and put into code. 

The use of a widely adopted deep learning framework: PyTorch, and benchmark datasets, like CIFAR-10, and CIFAR-100 made it easier to set up the experiments and compare our results with the original study. 

\subsection{What was difficult}
% List part of the reproduction study that took more time than you anticipated or you felt were difficult. 

% Be careful to put your discussion in context. For example, don't say "the maths was difficult to follow", say "the math requires advanced knowledge of calculus to follow". 
\subsubsection{Reproducing Table 01}

Despite the availability of the source code, we encountered a few issues during the reproducibility process. Initially, we faced a compatibility issue with different PyTorch versions, which required us to trial and error to resolve the issues across multiple versions. Furthermore, we encountered specific errors related to module imports and undefined models, which was resolved by importing the necessary models into the Python main script. Another challenge was the computational requirements for extensive experimentation across multiple architectures when working with the ImageNet dataset. This necessitated the use of internal GPU resources, as the Google Colab free tier had limitations in terms of GPU capabilities and runtime.

\subsubsection{Reproducing Table 02}
\kt{We experienced the following difficulties while reproducing the values of Table 02 in the original study:}
% \\Since \textit{imagenet/ILSVRC/Data/CLS-LOC/val} folder does not contain the sub-folders for each class, we had to prepare the dataset folder structure separately.

\begin{itemize}
    \item The ImageNet dataset was not available in the original source and it redirected to the \url{https://www.kaggle.com/c/imagenet-object-localization-challenge/data} website. We downloaded the dataset from the redirected site. The size of the dataset was relatively large ($\approx$ 160GB) and it took around 8 hours to download and unzip the dataset.
    \item The folder structure of the downloaded dataset was different from the dataset used in the original study. \textit{imagenet/ILSVRC/Data/CLS-LOC/val} folder did not contain the sub-folders for each class, so we had to prepare the dataset folder structure separately. 
    % The validation folder didn't have separate class folders and we explicitly re-arranged the validation folder by inserting the the data into class folders.\\
    \item We encountered a ``\textit{torch.cuda.OutOfMemory: RuntimeError}" due to the initial GPU memory we have been using. We had to transfer from 16GB GPU (NVIDIA A100) to 80 GB GPU (NVIDIA A100) memory in order to train ImageNet related 3 models.
\end{itemize}

\subsection{Communication with original authors}
% Document the extent of (or lack of) communication with the original authors. To make sure the reproducibility report is a fair assessment of the original research we recommend getting in touch with the original authors. You can ask authors specific questions, or if you don't have any questions you can send them the full report to get their feedback before it gets published. 

\kt{We tried contacting the original authors during our reproducibility study but we didn't receive any response. However, their well-documented code base and clear methodology description in the paper provided sufficient information for us to reproduce the results successfully.} 

\section{Conclusions}
Our study finds that the accuracy results reported in the original paper are reproducible within a threshold of 10\% with respect to the accuracy values reported in the original paper, and thus the authors' primary claim (network deconvolution improves the performance of deep learning models compared with batch normalization) is successfully verified. Network deconvolution emerges as a promising technique for improving model accuracy; although it may increase the training time by 2\% to 358\% depending on the architecture and training epochs.

% \newpage

% \subsection{Tables}
% \lipsum[12]
% See awesome Table~\ref{tab:table}.

% \begin{table}
%  \caption{Sample table title}
%   \centering
%   \begin{tabular}{lll}
%     \toprule
%     \multicolumn{2}{c}{Part}                   \\
%     \cmidrule(r){1-2}
%     Name     & Description     & Size ($\mu$m) \\
%     \midrule
%     Dendrite & Input terminal  & $\sim$100     \\
%     Axon     & Output terminal & $\sim$10      \\
%     Soma     & Cell body       & up to $10^6$  \\
%     \bottomrule
%   \end{tabular}
%   \label{tab:table}
% \end{table}

%Bibliography
\bibliographystyle{unsrt}  
\bibliography{references}  

\clearpage
% \section*{Appendix}
% Section 3.7: Issues encountered when associating components with milepost markers
% \begin{figure}[h]
% \vspace{-0.1cm}
%   \centering
%   % [width=\linewidth, frame]{figure-filename}
%     \setlength{\fboxsep}{0pt} % Adjust the padding
%     \setlength{\fboxrule}{0.5pt} % Adjust the border width
%   \fbox{\includegraphics[width=0.6\linewidth]{figures/test1234.jpg}}
%   \caption{Issues encountered when associating components with milepost markers}
%   \label{Appendix figure 1}
% \vspace{-0.3cm}
% \end{figure}

\end{document}